\pdfoutput=1

\documentclass[11pt]{article}

\usepackage[preprint]{acl}

\usepackage{times}
\usepackage{latexsym}

\usepackage[T1]{fontenc}

\usepackage[utf8]{inputenc}

\usepackage{microtype}

\usepackage{inconsolata}

\usepackage{graphicx}

\usepackage[utf8]{inputenc}
\usepackage{algorithm}
\usepackage{algpseudocode}
\usepackage{arydshln}
\usepackage{multirow}
\usepackage[colorinlistoftodos]{todonotes}
\usepackage{tcolorbox}
\usepackage{amsmath}
\usepackage{colortbl} 
\usepackage{subcaption}
\usepackage{booktabs}
\usepackage{colortbl} 
\definecolor{lightblue}{RGB}{173, 216, 230} 
\definecolor{softgreen}{RGB}{34, 139, 34}

\DeclareUnicodeCharacter{266B}{\textmusicalnote}

\title{RISE: Reasoning Enhancement via Iterative Self-Exploration in Multi-hop Question Answering}

\author{Bolei He\textsuperscript{\rm 1,2}\thanks{~~Equal contributions.} \quad Xinran He\textsuperscript{\rm 2}\footnotemark[1] \quad Mengke Chen\textsuperscript{\rm 2} \quad Xianwei Xue\textsuperscript{\rm 2} \\
\textbf{Ying Zhu}\textsuperscript{\rm 2} \quad \textbf{Zhen-Hua Ling}\textsuperscript{\rm 1}\thanks{~~Corresponding author.} \\
\textsuperscript{1}University of Science and Technology of China, Hefei, China \\
\textsuperscript{2}Baidu Inc., Beijing, China \\
\texttt{hebl@mail.ustc.edu.cn}, 
\texttt{zhling@ustc.edu.cn}, \\
\texttt{\{hexinran, xuexianwei, zhuying11\}@baidu.com}, \\
\texttt{anthreebody@gmail.com}  ` \\
}

\begin{document}
\maketitle
\begin{abstract}

Large Language Models (LLMs) excel in many areas but continue to face challenges with complex reasoning tasks, such as Multi-Hop Question Answering (MHQA). MHQA requires integrating evidence from diverse sources while managing intricate logical dependencies, often leads to errors in reasoning. Retrieval-Augmented Generation (RAG), widely employed in MHQA tasks, faces challenges in effectively filtering noisy data and retrieving all necessary evidence, thereby limiting its effectiveness in addressing MHQA challenges. To address these challenges, we propose \textbf{RISE:} \textbf{\underline{R}easoning Enhancement via \underline{I}terative \underline{S}elf-\underline{E}xploration}, a novel framework designed to enhance models’ reasoning capability through iterative self-exploration. Specifically, RISE involves three key steps in addressing MHQA tasks: question decomposition, retrieve-then-read, and self-critique. By leveraging continuous self-exploration, RISE identifies accurate reasoning paths, iteratively self-improving the model’s capability to integrate evidence, maintain logical consistency, and enhance performance in MHQA tasks. Extensive experiments on multiple MHQA benchmarks demonstrate that RISE significantly improves reasoning accuracy and task performance.



\end{abstract}

\section{Introduction}

\begin{figure}[h]
    \centering
    \includegraphics[width=0.46\textwidth]{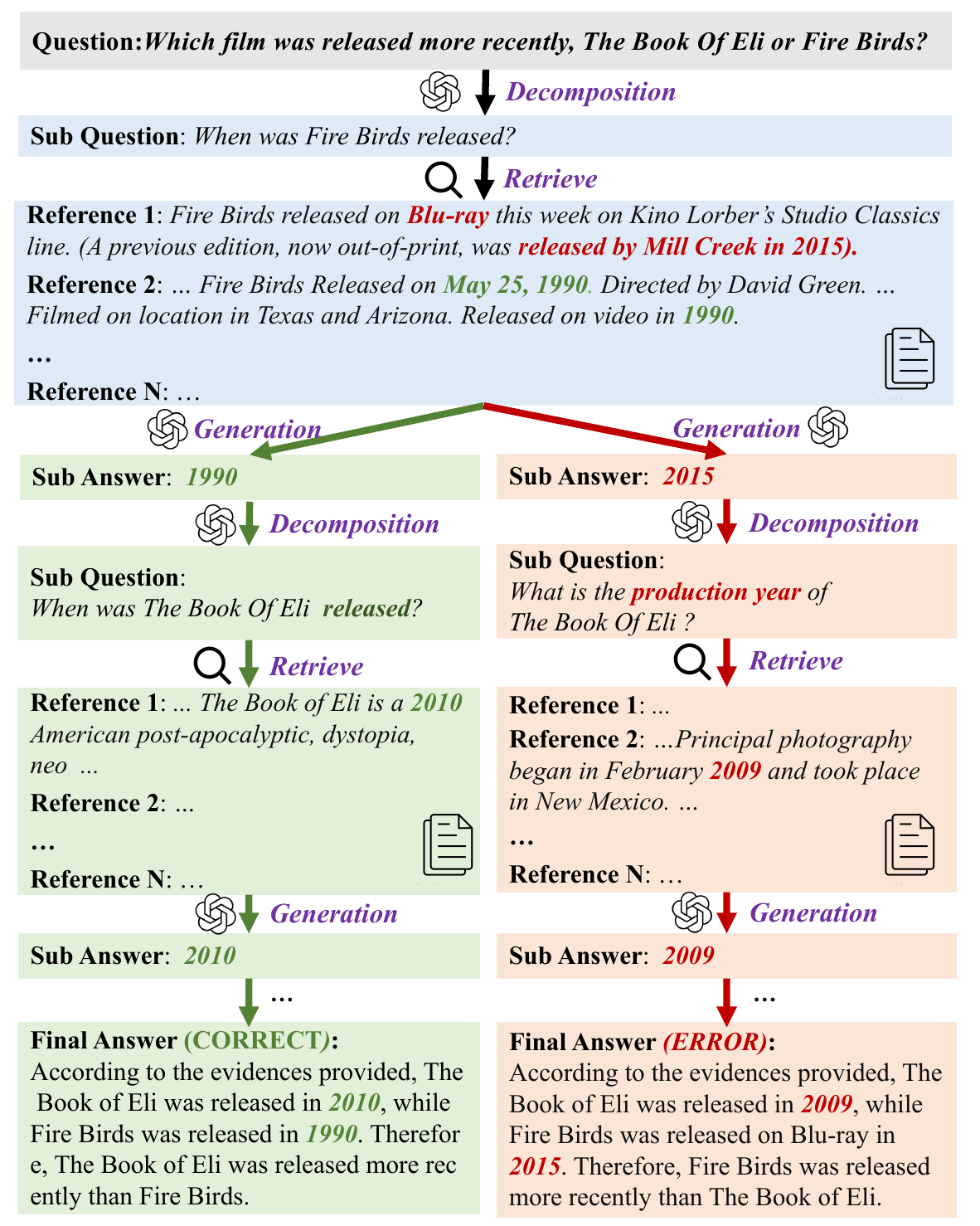}
    \caption{The upper part of the figure (\textcolor{lightblue}{blue}) illustrates an Evidence Aggregation Error, where the Blu-ray release year of \textit{Fire Birds} (\textcolor{red}{2015}) is mistaken for its theatrical release year. The lower part (\textcolor{softgreen}{green} and \textcolor{red}{red}) shows a Reasoning Decomposition Error. The incorrect path formulates the sub-question as the production year of \textit{The Book of Eli} (\textcolor{red}{2009}) instead of its release year (\textcolor{softgreen}{2010}).}
    \label{fig:intro}
\end{figure}


Large language models (LLMs) demonstrate outstanding capabilities in natural language understanding and generation~\cite{NEURIPS2020_1457c0d6,zhang2022opt,zeng2022glm,chowdhery2023palm,touvron2023llama}. However, LLMs still face challenges with complex Multi-Hop Question Answering (MHQA) tasks. MHQA requires models to integrate evidence from multiple sources and manage intricate logical relationships. This involves both retrieving and combining various pieces of evidence and constructing coherent reasoning chains. Prompt-based methods, such as Chain-of-Thought (CoT)~\cite{wei2023chainofthoughtpromptingelicitsreasoning,wang2023selfconsistencyimproveschainthought,yu2023generateretrievelargelanguage}, are employed to address MHQA by split complex problems into smaller, thereby harnessing the reasoning potential of LLMs. However, these methods often lack external knowledge, resulting in key evidence being overlooked and generate hallucinations~\cite{rawte2023survey, ji2023survey, ye2023cognitive}.


Retrieval-Augmented Generation (RAG) methods~\cite{guu2020retrieval, lewis2020retrieval,izacard2022few,nakano2021webgpt,asai2023selfrag, ma-etal-2023-query,yu2024rankragunifyingcontextranking,shi2024generatethengroundretrievalaugmentedgenerationmultihop} have been proposed to address the aforementioned challenges. By incorporating external knowledge, RAG effectively mitigates hallucination phenomena and achieves significant results in MHQA tasks through multiple retrievals. However, RAG is constrained by the performance of the retrievers, inevitably introducing noise. Additionally, the multi-round retrieval process may lead to error propagation, resulting in two main types of errors: \textbf{Evidence Aggregation Errors} and \textbf{Reasoning Decomposition Errors}. As illustrated in Figure~\ref{fig:intro}, Evidence Aggregation Errors occur when the model fails to accurately integrate evidence from multiple evidences, leading to hallucinations. Reasoning Decomposition Errors arise when problem decomposition phase generates sub-questions that do not align with original question's intent. These issues are particularly pronounced in smaller models with weaker reasoning capabilities.


Distillation and fine-tuning~\cite{uesato2022solvingmathwordproblems,luo2023wizardmathempoweringmathematicalreasoning,shridhar2023distillingreasoningcapabilitiessmaller} effectively enhance the reasoning capabilities of LLMs by leveraging large-scale models or high-quality, manually annotated data to improve performance. However, biases brought by human subjective annotations may undermine the performance of fine-tuning~\cite{casper2023openproblemsfundamentallimitations,lightman2023letsverifystepstep}, and these methods are costly, requiring substantial human or computational resources. Meanwhile, self-iteration methods~\cite{yuan2024selfrewardinglanguagemodels,wang2024selftrainingdirectpreferenceoptimization,madaan2023selfrefineiterativerefinementselffeedback} demonstrate tremendous potential in complex reasoning tasks. Unlike approaches that depend on large-scale models and manual annotations, self-iteration methods enable models to generate and learn from their own data, achieving outstanding results in complex tasks such as code generation and intelligent agents~\cite{jiang2023selfevolvecodeevolutionframework,ni2024nextteachinglargelanguage,qiao2024autoactautomaticagentlearning}. Nevertheless, research on combination self-iteration methods with RAG remains limited. The integration of these two approaches has the potential to improve performance in complex reasoning tasks and leads to cost reduction.



In this paper, we introduce an innovative framework, \textbf{RISE} (\textbf{\underline{R}}easoning Enhancement via \textbf{\underline{I}}terative \textbf{\underline{S}}elf-\textbf{\underline{E}}xploration), which combines the paradigms of RAG and self-iteration to address key challenges in MHQA tasks. Specifically, RISE defines three core actions: question decomposition, retrieve-then-read, and self-critique. By repeatedly executing these actions, the model autonomously explores accurate reasoning paths for problems. During this process, RISE accumulates experience datasets for the three actions and updates the model based on this experience. Through multiple iterations, RISE significantly enhances the model's reasoning capabilities in MHQA tasks. Experimental results demonstrate that RISE outperforms baseline methods on several MHQA benchmark datasets, strongly validating its effectiveness in solving MHQA tasks while offering lower usage costs. Our main contributions are as follows:




\begin{itemize}
    \item We propose RISE, which combines RAG and self-iteration to address two key challenges in MHQA tasks: Evidence Aggregation Errors and Reasoning Decomposition Errors.
    \item We design self-exploration mechanism, converts MHQA in RAG into multi-objective optimization problem, thus improving  model's reasoning capability and reducing costs.
    \item We integrate self-iteration paradigm with RAG, bridging gap in applying self-iteration strategies within MHQA RAG framework.

\end{itemize}


\begin{figure*}[htbp]
    \centering
    \includegraphics[width=0.96\textwidth]{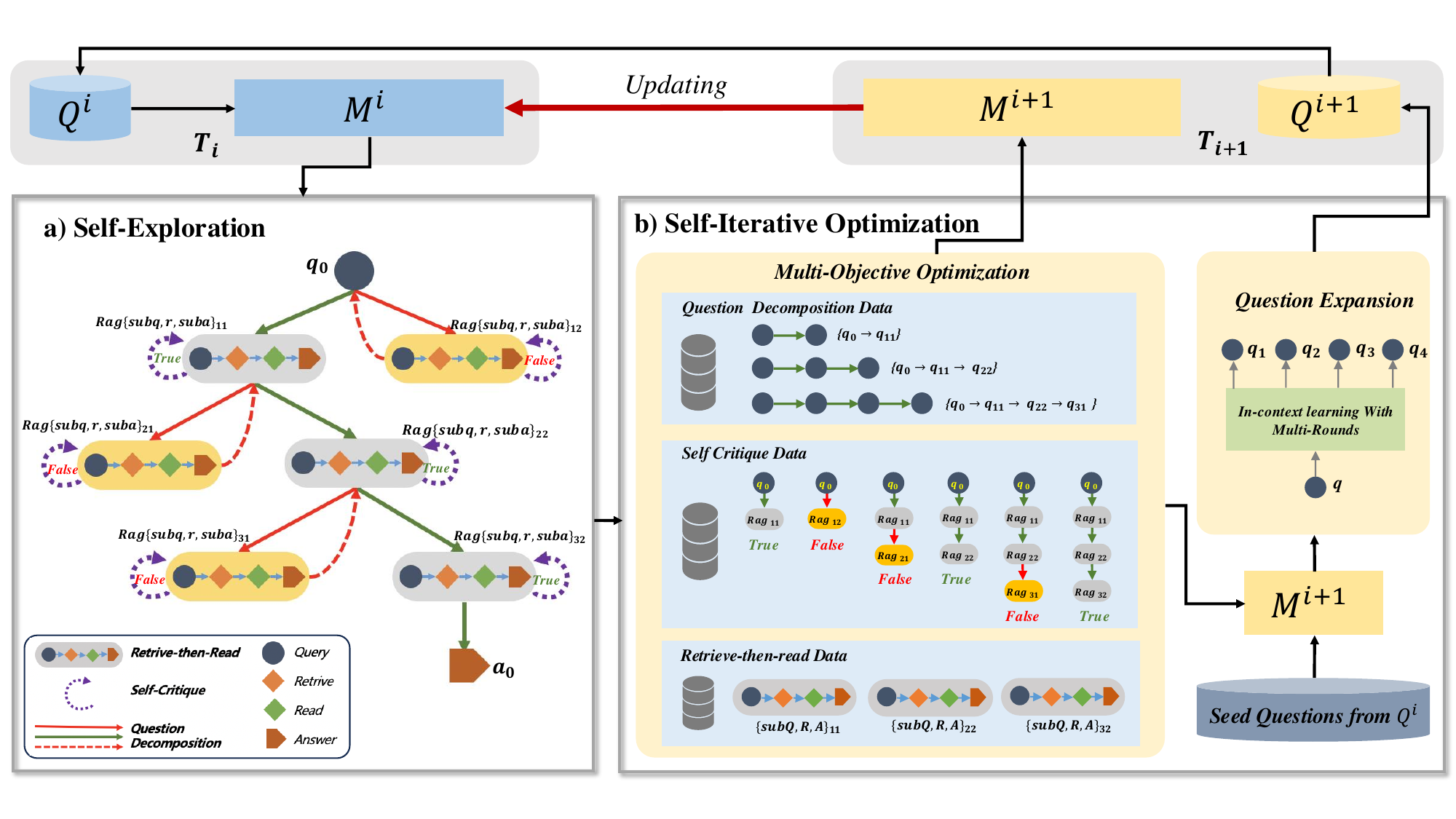}
    \caption{A complete iteration cycle in RISE. \textit{a) Self-Exploration}: Model $M^i$ decomposes complex questions $q_0$ into simpler sub-questions, generates sub-answers via retrieve-then-read, and evaluates their validity, leading to a final answer $a_0$. Interactions are stored as historical data $\mathcal{D}$. \textit{b) Iterative Optimization}: RISE optimizes model $M^i$ using historical data $\mathcal{D}$ to create an enhanced model $M^{i+1}$, which generates new questions $Q^{i+1}$ for the next cycle.}
    \label{fig:overview}
\end{figure*}

\section{Methods}

\begin{algorithm}[t]
    \caption{RISE}
    \label{alg:self-iterative}
    \textbf{Input:} Seed question set $\mathcal{Q}_0$, Initial model $M_0$, Retriever $R$, Maximum nodes $N_{max} = 20$
    \begin{algorithmic}[1]
        \State \textbf{Initialize:} History $\mathcal{H} \leftarrow \emptyset$, Model index $i \leftarrow 0$
        \While {True}
            \For {each question $q \in \mathcal{Q}_i$} 
               \State $n \leftarrow 0$ \Comment{\textcolor{red}{Start self-exploration.}}
                \While {$M_i(q, \mathcal{H}) =$ More information needed \textbf{and} $n < N_{max}$}
                    \State $subq \leftarrow M_i(\mathcal{H})$
                    \State $r \leftarrow R(subq)$
                    \State $suba \leftarrow M_i(subq, r)$
                    \State $\sigma \leftarrow M_i(subq, suba)$
                    \If {$\sigma = 1$}
                        \State Add $(subq, suba)$ to $\mathcal{H}$
                    \EndIf
                    \State $n \leftarrow n + 1$
                \EndWhile
                \State $a \leftarrow M_i(\mathcal{H})$
                \State $\sigma_{final} \leftarrow M_i(a, q, \mathcal{H})$
            \EndFor \Comment{\textcolor{red}{End self-exploration.}}
            \State $M_{i+1} \leftarrow \text{Multi-Objective Train}(M_i, \mathcal{H})$
            \State $\mathcal{Q}_{i+1} \leftarrow \text{Qustion Expansion}(M_{i+1}, \mathcal{Q}_i)$
            \State $i \leftarrow i + 1$
        \EndWhile
    \end{algorithmic}
    \noindent\textbf{Output:} Final model $M_i$
\end{algorithm}

\subsection{Overview}




In this section, we provide a concise description of \textbf{RISE}, focusing on its algorithmic process. As shown in algorithm~\ref{alg:self-iterative}, RISE begins with a seed question set $\mathcal{Q}_0$ and an initial model $M_0$. The model iteratively performs self-exploration for each question $q \in \mathcal{Q}_i$, with details presented in Section~\ref{Self-Exploration Mechanism}. The exploration results are stored as historical experience $\mathcal{H}$. After completing the exploration for all questions, the accumulated experiences optimize the model through multi-objective training, yielding an enhanced model $M_{i+1}$. Subsequently, $M_{i+1}$ expands the question set based on the previous seed questions $\mathcal{Q}_i$, generating $\mathcal{Q}_{i+1}$ to initiate the next round of exploration. This self-iterative process enables RISE to continuously improve capabilities without external supervision, leveraging the model’s intrinsic potential.

\subsection{Self-Exploration Mechanism}
\label{Self-Exploration Mechanism}


The self-exploration mechanism enables the model to address complex problems through iterative reasoning, comprising three core actions: question decomposition, retrieve-then-read, and self-critique. These actions collectively form a structured exploration pathway, with the resulting information collected as historical data to support the model’s self-improvement in complex problem-solving. The related prompts are provided in Appendix \ref{appendix:self-exploration}.

\noindent \textbf{Question Decomposition.} In this task, the model incrementally decomposes the initial complex question into fine-grained sub-questions. At the $\mathit{t}$-th exploration node, the model uses previously explored sub-questions and answers as historical information, denoted as $\mathcal{H} = {(subq_1, suba_1), \cdots, (subq_{t-1}, suba_{t-1})}$. The original question $q_0$ is combined with $\mathcal{H}$ and input into model $M$ to generate the next sub-question. 
The model ends exploration by generating the final answer if the historical information suffices for $q_0$. Formally, this process is represented as Formula \ref{decomposition}:

\begin{align}
    \label{decomposition}
    subq_{t}&=\mathcal{F}_{d}(M,\mathcal{H} , q_0)  \\
    a_0 &= M(q_0, \mathcal{H}), \quad \text{if } \mathcal{H} \text{ is sufficient.}
\end{align}

Additionally, all decomposition steps, including the original question and generated sub-questions, are recorded to form the dataset $\mathcal{D}_{\text{d}} = \left\{\left\{q_0, \mathcal{H}, subq\right\}_{i=1}^{n_{\text{p}}}\right\}^{N_{\text{q}}}$. By leveraging this fine-grained and structured dataset, the model learns the logical dependencies and relationships between questions and sub-questions, thereby improving its ability to decompose complex problems.

\noindent \textbf{Retrieve-then-Read.} This task follows the standard RAG paradigm to provide evidence-based answers for sub-questions. At the $\mathit{t}$-th exploration node, a retriever obtains relevant fragments $r_t$ based on the sub-question, and model $M$ generates answer using the retrieved evidence:

\begin{align}
suba_{t} = \mathcal{F}_{g}(M, subq_t^i, r_t)
\label{gen}
\end{align}

Each sub-question and its answer form an exploration node $(subq_i, suba_i)$, added to the historical information $\mathcal{H}_{t+1} = \mathcal{H}_t \cup \{(subq_i, suba_i)\}$. All nodes are recorded to construct the dataset $\mathcal{D}_{\text{r}} = \left\{\left(subq, r, suba\right)_{i=1}^{n_{\text{p}}}\right\}^{N_{\text{q}}}$. Training on this dataset helps the model integrate evidence into reasoning, improving answer accuracy and reliability.


\noindent \textbf{Self-Critique.} In this task, the model’s critique capability is incorporated into the exploration process. Specifically, after completing the question decomposition and retrieve-then-read tasks at the $\mathit{t}$-th exploration node, the model $M$ critiques the relevance and utility of the node for solving the original question and outputs a binary decision. If critiqued as \text{True}, it is retained, and exploration proceeds to the next step. If critiqued as \text{False}, the node is temporarily stored, and the process reverts to the preceding valid node to generate a new node. This process is formalized in Formula \ref{Critique}:

\begin{align}
    \label{Critique}
    \sigma_t = \mathcal{F}_{\text{c}}(M, subq_t, suba_t), \quad \sigma_t \in \{0, 1\}
\end{align}


Similarly, we record critique historical information and then construct the dataset $\mathcal{D}{c} = \left\{\left\{\langle subq, suba \rangle, \sigma\right\}_{i=1}^{n_{\text{p}}}\right\}^{N_{\text{q}}}$, to enhance the self-critique capabilities of the model, ensuring logical consistency and relevance within the exploration.




%

\subsection{Self-Iterative Optimization}

RISE is a self-iterative fine-tuning framework that optimizes the model in each training round based on data generated by the model itself, gradually enhancing its generalization ability and reasoning performance. Through a closed-loop iteration of data generation and model training, RISE effectively uncovers the model’s reasoning potential in complex tasks, driving continuous self-improvement.

\noindent\textbf{Initialization.} We initially use randomly sampled question set $\mathcal{Q}_0$ from the training sets of three tasks—2Wiki, HotpotQA, and Musique, with 800 examples from each. Subsequently, we employ self-exploration mechanism to automatically expand and collect the three types of dataset. Subsequently, we employ self-exploration mechanism to automatically expand and collect $\mathcal{D}_d$, $\mathcal{D}_r$, and $\mathcal{D}_c$ datasets for subsequent model training.

\noindent\textbf{Multi-Objective Optimization.}
These three datasets, $\mathcal{D}{d}$, $\mathcal{D}{r}$, and $\mathcal{D}_{c}$, are interconnected, with sample sizes ranging from 2k to 8k (detailed statistics are provided in Appendix~\ref{tab:train_data}). We believe that joint training facilitates complementary learning and enhances model capabilities. Therefore, we adopt a multi-objective optimization approach to integrate the objectives of different tasks into a unified optimization goal. The effectiveness of this approach is validated through ablation study at Section~\ref{sec:ablation}. The overall loss function is defined as follows Formula \ref{formula:1}:


\begin{align}
    \mathcal{L} = \alpha\mathcal{L}_{d} +   \beta\mathcal{L}_{r} +  \gamma \mathcal{L}_{c} 
    \label{formula:1}
\end{align}

\begin{table}[t]
\centering
\small
\begin{tabular}{c|cccccccccc}
\hline
         & \multicolumn{5}{c}{weights ratios} \\ \hline
$\alpha$ & 1 & 1 & 4 & 1 & 1 \\ 
$\beta$  & 1 & 4 & 1 & 2 & 3 \\ 
$\gamma$ & 4 & 1 & 1 & 3 & 2 \\ 
Accuracy & 41.55 & 40.32 & 41.94 & 38.71 & 37.63 \\ \hline
$\alpha$ & 2 & 2 & 3 & 3 & 2 \\
$\beta$  & 1 & 3 & 1 & 2 & 2 \\ 
$\gamma$ & 3 & 1 & 2 & 1 & 2 \\
Accuracy & 39.78 & 40.32 & 43.01 & 41.13 & 44.27 \\
\hline
\end{tabular}
\caption{Performance on the 2WikiMultiHopQA dataset under varying weight ratios of $\alpha$, $\beta$, and $\gamma$.}
\label{tab:transposed_loss_function_results}
\end{table}




Here, $\alpha$, $\beta$, and $\gamma$ represent task weights, determined by the proportion of each capability in the training data. Specifically, $\mathcal{L}_d$ (Formula~\ref{formula:2}) models the autoregressive loss for sub-question generation, $\mathcal{L}_r$ (Formula~\ref{formula:3}) models the loss for sub-answer generation conditioned on retrieved context, and $\mathcal{L}_c$ (Formula~\ref{formula:4}) denotes a binary classification loss for self-critique judgments, where True and False represent the predicted probabilities of the respective tokens.

\footnotetext{$P(True) = \frac{P(Token(True))}{P(Token(True)) + P(Token(False))}$ and $P(False)$ accordingly.}


\begin{align}
    \mathcal{L}_d &= -\sum_{i} \log P(\text{subq}_i \mid q_0, \text{subq}_{<i}) \label{formula:2} \\
    \mathcal{L}_r &= -\sum_{i} \log P(\text{suba}_i \mid \text{subq}_i, r_i) \label{formula:3} \\
    \mathcal{L}_c &= -\sigma \log P(True \mid \text{subq}_i, \text{suba}_i) \notag \\
                 &\quad - (1 - \sigma) \log P(False \mid \text{subq}_i, \text{suba}_i) \label{formula:4}
\end{align}

Meanwhile,  Experimental results (Table~\ref{tab:transposed_loss_function_results}) indicate that the weights assigned to different tasks have an impact on model performance, and appropriate weight adjustments facilitate fine-grained performance optimization. Notably, to avoid potential overfitting caused by manual weight tuning which may affect the final evaluation we do not perform any fine-tuning of the task weights in our experiments. Instead, we adopt a uniform weighting strategy, assigning equal weights to all tasks.

\noindent\textbf{Question Expansion}
After completing multi-objective optimization, we use the questions generated in the previous iteration as seed data for $M^{i+1}$ to perform question expansion, thereby acquiring training data for the next iteration. This method is inspired by \cite{wang2023self}, leveraging multi-round in-context learning to ensure the diversity and richness of the newly generated questions. Detailed information about the question expansion prompts is provided in the appendix Figure~\ref{appendix:multi-hop-prompt}.


\section{Experiments Setup}


\noindent \textbf{Datasets:} 
For the main experiments, we use three QA datasets: 2WikiMultiHopQA (2WIKI)~\cite{ho-etal-2020-constructing}, HotpotQA (Hotpot)~\cite{yang-etal-2018-hotpotqa}, and MuSiQue (MSQ)~\cite{trivedi-etal-2022-musique}, which provide diverse reasoning challenges to evaluate the robustness of our framework. Additionally, for the analysis experiments, we include Natural Questions (NQ)~\cite{kwiatkowski2019natural}, Web Questions (WebQ)~\cite{berant2013semantic} and TriviaQA~\cite{joshi2017triviaqa} to assess the model’s performance on open-domain Question Answering tasks, further extending the evaluation scope.

\noindent \textbf{Models and Methods:} 
In our experiments, we use LLaMA-3.1-8B~\cite{dubey2024llama} as the base model for our method in main experiments. Similarly, most of the reproduced methods are also implemented using LLaMA-3.1-8B. Additionally, based on the characteristics of MHQA tasks, we select and reproduce a variety of methods, categorized into non-retrieval-based methods and retrieval-based methods. Non-retrieval-based methods include Naive LLM (LLaMA-3.1-8B, GPT-3.5-turbo), CoT~\cite{wei2023chainofthoughtpromptingelicitsreasoning}, CoT-SC~\cite{wang2023selfconsistencyimproveschainthought} and GenRead~\cite{yu2023generateretrievelargelanguage}, while the retrieval-based methods consist of Naive RAG, Self-Ask~\cite{press2023measuringnarrowingcompositionalitygap}, WebGLM~\cite{liu2023webglm}, Self-RAG~\cite{asai2023selfrag}, RRR~\cite{ma-etal-2023-query}, and GenGround~\cite{shi2024generatethengroundretrievalaugmentedgenerationmultihop}. In the analysis experiments, we employ GPT-4o\footnote{We use GPT models accessed via the OpenAI API: \url{https://openai.com/api/}.} as the evaluation model, combining subjective analysis with specific metrics to comprehensively assess model performance.

\noindent \textbf{Retrieval:} 
We adopt a two-stage retrieval framework \cite{liu2023webglm}, consisting of coarse-grained web search (via Chrome) followed by fine-grained LLM-enhanced retrieval. We consistently use the same retrieval method to reproduce results for other approaches that incorporate retrievers.


\noindent \textbf{Evaluation Metrics:} 
We assess model performance primarily using standard metrics for question answering tasks: Accuracy (Acc), F1 score (F1), and Exact Match (EM), which together provide a comprehensive measure of answer correctness and completeness. In addition to answer quality, we evaluate the generated reasoning chains by examining their length as an objective measure of complexity. Furthermore, we conduct a qualitative assessment of the reasoning chains from four subjective perspectives: conciseness, rationality, sequencing, and goal orientation.

We provide comprehensive experimental details in Appendix~\ref{experiment details}, including implementation details, datasets, and other relevant information.

\begin{table*}[]
\centering
\scalebox{0.85} {
\begin{tabular}
{p{4.5cm}p{2.5cm}|p{1.35cm}p{1.35cm}p{1.35cm}|p{1.35cm}p{1.35cm}p{1.35cm}p{1.35cm}llcccccc}
\hline
\multirow{2}{*}{\textbf{Method}} &
  \multirow{2}{*}{\textbf{Model}} &
  \multicolumn{3}{c|}{\textbf{MHQA}} &
  \multicolumn{3}{c}{\textbf{SHQA}} \\
   & &
  \textbf{2WIKI} &
  \textbf{Hotpot} &
  \textbf{MSQ} &
  \multicolumn{1}{c}{\textbf{NQ}} &
  \multicolumn{1}{c}{\textbf{WebQ}} &
  \multicolumn{1}{c}{\textbf{Trival}} \\ \hline
\rowcolor[gray]{0.95} \multicolumn{8}{c}{\textit{w/o retrieval}} \\ 
\rowcolor{green!10} Naive LLM                                                                       & LLaMA-3.1-8B    & 35.90 & 27.30 & 11.30 & 57.50 & 61.25  & 71.50 \\
\rowcolor{green!10}                                                                                 & GPT-3.5-turbo   & 47.10 & 41.50 & 19.10 & 57.25 & 58.30  & 80.25 \\
\rowcolor{green!10} CoT~\cite{wei2023chainofthoughtpromptingelicitsreasoning}                       & LLaMA-3.1-8B    & 43.00 & 34.60 & 16.20 & 56.75 & 62.00  & 71.75 \\
\rowcolor{green!10} CoT-SC\textsuperscript{*}~\cite{wang2023selfconsistencyimproveschainthought}    & GPT-3.5-turbo   & 21.00 & 30.60 & 8.90  & 39.75 & 38.50  & 79.25 \\
\rowcolor{green!10} GenRead~\cite{yu2023generateretrievelargelanguage}                              & LLaMA-3.1-8B    & 20.00 & 28.40 & 10.30 & 46.25 & 43.00  & 67.00 \\
\rowcolor[gray]{0.95} \multicolumn{8}{c}{\textit{w retrieval}} \\ 
\rowcolor{green!10} Naive RAG                                                                       & LLaMA-3.1-8B    & 43.50 & 37.50 & 15.00 & 51.75 & 57.75  & 73.75 \\
\rowcolor{green!10} Self-Ask~\cite{press2023measuringnarrowingcompositionalitygap}                  & LLaMA-3.1-8B    & 22.90 & 29.70 & 12.70 & 44.50 & 46.50  & 66.00 \\
\rowcolor{yellow!10} WebGLM~\cite{liu2023webglm}                                                    & LLaMA-3.1-8B    & 37.60 & 36.50 & 13.30 & 55.75 & 62.25  & 73.75 \\
\rowcolor{yellow!10} Self-RAG\textsuperscript{*}~\cite{asai2023selfrag}                             & LLaMA2-7B       & 32.00 & 30.20 & 8.00  & 54.50 & 60.25  & 74.25 \\
\rowcolor{yellow!10}                                                                                & LLaMA2-13B      & 30.80 & 29.40 & 8.80  & 55.25 & 58.75  & 75.00 \\
\rowcolor{yellow!10} RRR~\cite{ma-etal-2023-query}                                                  & LLaMA-3.1-8B    & 23.70 & 11.80 & 5.40  & 22.50 & 30.50  & 36.75 \\
\rowcolor{green!10}                                                                                 & GPT-3.5-turbo   & 28.20 & 29.70 & 8.70  & 57.75 & 56.50  & 80.50 \\
\rowcolor{yellow!10} GenGround~\cite{shi2024generatethengroundretrievalaugmentedgenerationmultihop} & LLaMA-3.1-8B    & 37.90 & 36.10 & 17.80 & 48.50 & 44.50  & 75.25 \\ \hline
\rowcolor{yellow!10} \textbf{RISE(Ours)}                                                            & LLaMA-3.1-8B    & 49.40 & 40.50 & 21.70 & 59.50 & 62.50  & 80.25 \\ \hline
\end{tabular}
}

\caption{Comparison of RISE’s accuracy with other methods on 2WikiMultiHopQA, HotpotQA, MuSiQue, Natural Questions, Web Questions, and TriviaQA. Methods marked with an asterisk (*) involve specific considerations: CoT-SC uses GPT-3.5 due to LLaMA-3.1’s instruction-following limitations, and Self-RAG employs public model weights as its dataset is unavailable. Other methods use LLaMA-3.1-8B. \textbf{\textcolor{green!20}{\rule{3mm}{3mm}}} denote \textit{Prompting-based Methods}, while \textbf{\textcolor{yellow!20}{\rule{3mm}{3mm}}} denote \textit{Finetuning-based Methods}. Due to space constraints,\textcolor{red}{\textbf{ \textit{F1}}} and \textcolor{red}{\textbf{\textit{EM}}} metrics are in Appendix~\ref{sec:appendix-metrics}.}
\label{tab:main_result}
\end{table*}

\begin{figure}[t]
    \centering
    \includegraphics[width=0.93\linewidth]{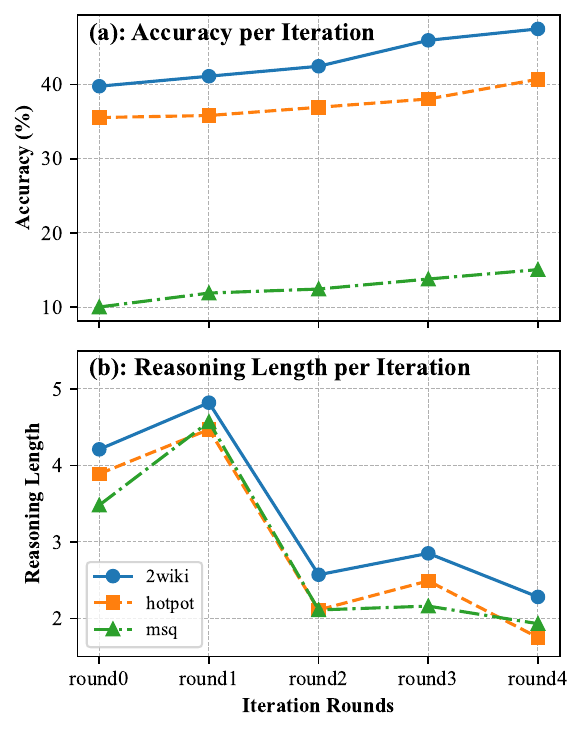}
    \caption{Changes in model accuracy (a) and reasoning length (b) across datasets. Accuracy consistently improves across datasets, while reasoning length, despite some fluctuations, shows an overall decreasing trend.}
    \label{fig:line_chart}
\end{figure}

\section{Results and Analysis}



In this section, we evaluate RISE from three aspects. First, we validate effectiveness of multiround self-iterative and compare RISE with mainstream MHQA methods. Second, we conduct an in-depth analysis of the performance of question decomposition, retrieve-then-read, and self-critique using objective metrics and AI-based evaluations. Finally, we conduct ablation studies to verify the importance of different tasks in enhancing performance.


\subsection{Overall Performance}


\noindent \textbf{RISE Outperforms Other Methods:}
Table \ref{tab:main_result} presents the experimental results across three MHQA datasets and three SHQA datasets. We observe that retrieval-based enhancement is crucial for MHQA tasks. While CoT achieves relatively good performance, other non-retrieval methods generally perform worse than most RAG approaches with the same model. For the relatively simpler SHQA tasks, retrieval-based enhancement does not seem to offer significant advantages. Notably, RISE achieves outstanding results in both task types over all datasets, even surpassing GPT-3.5. Furthermore, our method excels in F1 and EM metrics, demonstrating its efficiency (additional metrics are provided in Appendix~\ref{sec:appendix-metrics}).

\begin{figure*}[h]
    \centering
    \includegraphics[width=0.93\linewidth]{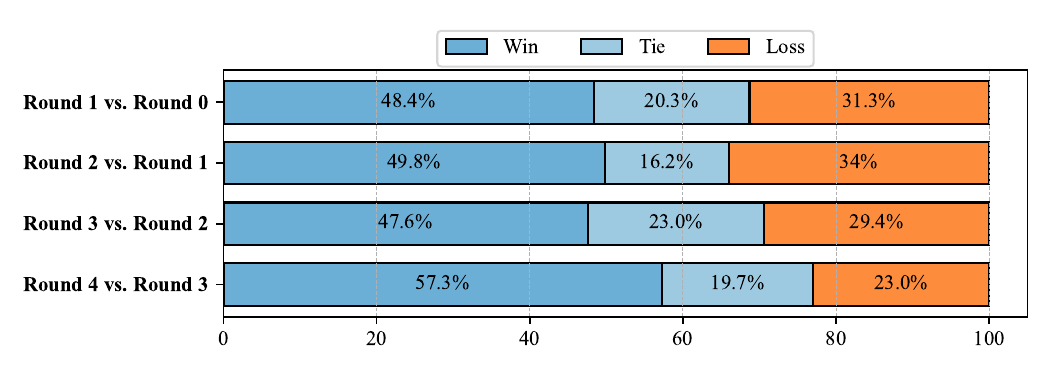}
    \caption{Evaluating the win rates between the current and previous iterations using GPT-4o to assess model’s question decomposition capability. Results indicate that each new iteration consistently outperforms the previous one in subjective effectiveness, demonstrating RISE’s continuously enhance the model’s decomposition capability.}
    \label{fig:win_tie_loss}
\end{figure*}

\begin{figure*}[h]
    \centering
    \includegraphics[width=0.95\linewidth]{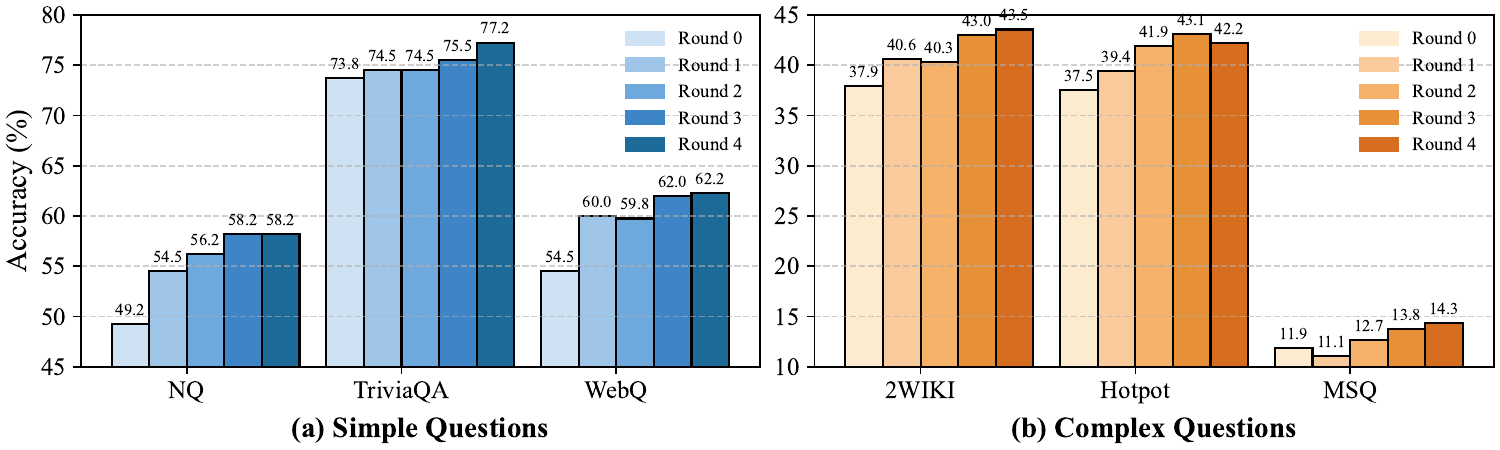}
    \caption{Changes in the model’s retrieve-then-read capability. (a) Results on simpler datasets (NQ, TriviaQA, WebQ), (b) Results on more complex datasets (2Wiki, HotpotQA, MSQ), where accuracy shows consistent growth with each iteration, even in challenging scenarios.}
    \label{fig:summarization_bar}
\end{figure*}

\noindent \textbf{Steady Performance Improvement:} Meanwhile, as shown in Figure \ref{fig:line_chart} \textit{(a) Accuracy per Iteration}, we illustrate how the model's accuracy evolves over four iterations on multiple datasets. The results demonstrate a consistent upward trend in accuracy with each iteration, further validating the effectiveness of our proposed self-training method in improving the model's overall performance.

\subsection{Analysis Experiments}


\subsubsection{Question Decomposition Capability}

To evaluate improvement in the model's decomposition capability for MHQA tasks, we first analyze the changes in reasoning length. As shown in Figure \ref{fig:line_chart} \textit{(b) Reasoning Length per Iteration}, accuracy steadily improves, while reasoning length initially increases and then decreases, ultimately showing downward trend. This trend reflects model’s decomposition ability progressively improves over iterations.

To further analyze changes in decomposition ability, we using GPT-4o as a judge to evaluate the model’s query decomposition across four dimensions (including conciseness, rationality, sequencing and goal orientation, see Appendix~\ref{appendix:gpt4decomposition} for more details.). As illustrated in Figure \ref{fig:win_tie_loss}, we compare the performance of the model across iterations and observe newer model consistently outperforms the previous iteration. These findings demonstrate that self-training not only improves reasoning paths but also enhances the rationality of decomposition.

\subsubsection{Retrieve-then-Read Capability}
In MHQA tasks, models often struggle to integrate logical information from extensive evidence, especially in filtering irrelevant content. To evaluate the changes in the model’s summarization capability over iterations, we disable the decomposition functionality and instead allow model to perform single-round retrieval and direct question-answering. To ensure robustness in the experiments, we introduce simpler datasets such as NQ, WebQ, and TriviaQA (Figure \ref{fig:summarization_bar} \textit{(a) Simple Questions}) while retaining the complex datasets from main experiments (Figure \ref{fig:summarization_bar} \textit{(b) Complex Questions}).
The experimental results show that, as iterations progress, RISE consistently improves its performance across six datasets. This demonstrates the advantage of RISE in MHQA tasks and its effectiveness in conventional QA tasks, further validating its generalizability.




\subsubsection{Self-Critique Capability}
To evaluate the changes in the model’s self-critique capability, we designed a third set of experiments. In this experiment, both our model and GPT-4o assess the same set of decomposition results, with GPT-4o serving as a reference. By analyzing the consistency between our model and GPT-4o evaluations, we measure the improvement in the model’s self-critique capability. As shown in Table \ref{tab:criticism}, the consistency between our model and GPT-4o steadily increases with each iteration. This indicates that the iterative process in RISE effectively enhances the model’s self-criticism capability. (For more experiment details see Appendix~\ref{appendix:self-criticism}.)



\begin{table}[t]
\centering
\begin{tabular}{lccc}
\hline
\multirow{2}{*}{} & \multicolumn{3}{c}{Consistency with GPT-4o (\%)} \\ \cline{2-4}
                        & 2WIKI          & HotpotQA       & MSQ         \\ \hline
Round 1                 & 74.30          & 64.70          & 60.00       \\
Round 2                 & 72.67          & 66.30          & 76.00        \\
Round 3                 & 79.67 & 77.33 & \textbf{79.33} \\ 
Round 4                 &  \textbf{80.67}         & \textbf{79.33} & 78.00 \\ \hline

\end{tabular}
\caption{Consistency analysis with GPT-4o on each datasets. The results show progressive improvements in consistency with GPT-4o, highlighting the model's enhanced self-critique ability through iterative training.}
\label{tab:criticism}
\end{table}

\subsection{Ablation Study}
\label{sec:ablation}
To evaluate the impact of each synthesized training dataset on the model's performance, we conduct an ablation study. As shown in Table \ref{tab:ablation}, the experiment uses the same three MHQA datasets as before and the three training datasets generated in the round1, with accuracy as the primary evaluation metric.

Removing the question decomposition dataset leads to accuracy drop of 3.5\% on 2Wiki, highlighting its importance in enabling effective multi-hop reasoning. Excluding the retrieve-then-read dataset causes accuracy declines on HotpotQA (2.77\%) and Musique (2.43\%), highlighting the importance of this dataset in synthesizing evidence from diverse sources to mitigate the impact of noise. The removal of the self-critique dataset results in consistent accuracy reductions across all three datasets, emphasizing its pivotal function in refining reasoning paths processes. These results demonstrate the complementary and indispensable contributions of the question decomposition, retrieve-then-read, and self-critique datasets to the model’s performance.

Furthermore, we conduct separate training for the three tasks (Separate), where three LLMs are individually trained for decomposition, retrieve-then-read, and self-critique tasks. Compared to joint training (RISE), the accuracy of separate training is consistently lower across all datasets.



\begin{table}[t]
\centering
\begin{tabular}{llll}
\hline
\multirow{2}{*}{} & 2WIKI & Hotpot & MSQ \\
                        & Acc   & Acc      & Acc    \\ \hline
w/o Decomp       & 37.63 & 33.89    & 11.08          \\
w/o R-t-R          & 40.59 & 33.06    & 9.46          \\
w/o Critique            & 38.98 & 33.89    & 10.27     \\
Separate            & 40.86 &  34.72        &  10.54          \\
RISE                    & \textbf{41.13} & \textbf{35.83}    & \textbf{11.89}     \\ \hline
\end{tabular}
\caption{Ablation study on 2WIKI, HotpotQA, and MSQ, showing the impact of removing individual tasks (\textit{Question Decomposition}, \textit{Retrieve-then-Read}, and \textit{Self-Critique}) and comparing joint training (\textit{RISE}) with separate training (\textit{Separate}) of individual tasks.}

\label{tab:ablation}
\end{table}

\section{Related Works}
\noindent \textbf{Multi-hop Question Answering:} 
MHQA tasks address questions that require integrating information from multiple sources and performing multi-step reasoning to produce a complete answer~\cite{zhang-etal-2024-end,li-du-2023-leveraging}. Question decomposition has been a pivotal approach for understanding and solving multi-hop questions, some works~\cite{NEURIPS2022_9d560961,wang2023selfconsistency, zhou2023leasttomost,shi-etal-2024-generate} leverage LLMs to divide complex questions into simpler single-hop sub-questions that are solved sequentially. Self-Ask~\cite{press2023measuringnarrowingcompositionalitygap} uses LLMs to generate and resolve follow-up sub-questions with an external search engine. However, the effectiveness of these approaches depends on LLM’s inherent question decomposition capability, which constrained by hallucinations.

\noindent \textbf{Retrieval-Augmented Generation for MHQA:} 
RAG~\cite{guu2020retrieval, lewis2020retrieval,izacard2022few,nakano2021webgpt,asai2023selfrag, ma-etal-2023-query,yu2024rankragunifyingcontextranking,shi2024generatethengroundretrievalaugmentedgenerationmultihop} integrates retrieval with generation to solve knowledge-intensive tasks~\cite{zhu2024largelanguagemodelsinformation,10448015}. The original RAG framework excels at single-hop QA but faces significant challenges in handling multi-hop QA and complex reasoning tasks~\cite{lewis2020retrieval, 10.1145/3589334.3645363}. 

To address these challenges, various methods have been proposed. Chain of Thought (CoT)~\cite{wei2023chainofthoughtpromptingelicitsreasoning} and Tree of Thought (ToT)~\cite{yao2023treethoughtsdeliberateproblem} are integrated with RAG to enable multi-step reasoning and iterative retrieval~\cite{press2023measuringnarrowingcompositionalitygap,yao2023reactsynergizingreasoningacting,zhou2023leasttomost,khattab2023demonstratesearchpredictcomposingretrievallanguage}, allowing the model to incorporate a broader range of external knowledge and improve its reasoning capabilities. However, existing retrieval-augmented systems are inevitably affected by the limitations of retrievers, often introducing irrelevant or noisy information~\cite{yin-etal-2023-alcuna,10.1145/3589334.3645363,ma-etal-2023-query}.  Enhancing the model’s reasoning capabilities to filter noise and focus on critical evidence is essential for accurate summaries, which our method achieves through reasoning decomposition, improving both logical reasoning and QA performance.

\noindent \textbf{Self-Improvement in Large Language Models:} 
Self-improvement refers to the process by which models generate and utilize their own output data to enhance performance~\cite{zelikman2024selftaughtoptimizerstoprecursively,singh2024humandatascalingselftraining,gulcehre2023reinforcedselftrainingrestlanguage}. Existing approaches, such as self-training~\cite{du-etal-2021-self} and self-play~\cite{yuan2024selfrewardinglanguagemodels,chen2024selfplayfinetuningconvertsweak}, leverage pseudo-label generation and iterative policy optimization to improve the utilization of unlabeled data and enhance decision-making capabilities. Self-Rewarding~\cite{yuan2024selfrewardinglanguagemodels} employs the LLM-as-Judge paradigm to strengthen reasoning abilities, while Self-Refine~\cite{madaan2023selfrefineiterativerefinementselffeedback} iteratively optimizes generated outputs through self-feedback mechanisms.

In complex tasks like code generation and agent-based learning, self-improvement proves effective. Methods such as Self-Evolve~\cite{jiang2023selfevolvecodeevolutionframework}, NExT~\cite{ni2024nextteachinglargelanguage}, and AutoAct~\cite{qiao2024autoactautomaticagentlearning} leverage self-feedback, self-guided tracking, and self-planning to enhance performance. However, the application of self-iterative techniques in RAG scenarios remains underexplored. Our method addresses this gap by integrating self-exploration into RAG to generate diverse training data, enabling continuous model evolution and enhancing performance in complex tasks.

\section{Conclusion}
We propose RISE, a framework that addresses two key errors in MHQA tasks: Evidence Aggregation and Reasoning Decomposition. Through self-exploration, RISE continuously enhances reasoning capabilities. Additionally, RISE integrates self-iterative paradigm with RAG framework, bridging the gap in applying self-iterative strategies to MHQA scenarios without requiring manual intervention or reliance on large models, thereby offering a cost-effective solution. Experimental results on MHQA benchmarks demonstrate significant improvements in reasoning accuracy and task performance, highlighting RISE's robustness and adaptability in tackling complex reasoning challenges.



\section*{Limitation}
While RISE achieves strong performance in complex reasoning tasks, there remain opportunities for further enhancement. The current framework relies on external retrieval mechanisms without explicit optimization, which may limit the quality of evidence for downstream reasoning. Future work could explore self-improvement across the entire pipeline—spanning question decomposition, retrieval, generation, and reflection—to achieve more seamless integration and efficiency. 

\bibliography{custom}

\appendix
\clearpage

\section{Appendix}
\label{sec:appendix}

\subsection{Prompts}

\subsubsection{Self-Exploration Prompts}
\label{appendix:self-exploration}
We designed detailed prompts for the three tasks in the self-exploration phase: question decomposition (Figure~\ref{appendix:decomposition prompt}), retrieve-then-read(Figure~\ref{appendix:generation prompt}), and self-critique(Figure~\ref{appendix:critique prompt}). The examples used in the decomposition prompt are inspired by self-ask~\cite{press2023measuringnarrowingcompositionalitygap}.

\begin{figure*}[t]
\noindent
\begin{tcolorbox}[title = {Decomposition Prompt}] 
Instruction: Please answer the following questions according to the given format. Strictly follow each format specification, as this will ensure consistency and clarity in your response.
\\

- Only add follow-up questions if additional details are needed to arrive at the final answer.\\
- For each follow-up question, use exactly this format: 'Follow up: {question}'\\
- Ensure each follow-up question is direct and structured to be easily searchable, focusing on key information for efficient search engine retrieval.\\
- For each answer to a follow-up question, use exactly this format: 'Intermediate answer: {answer}'\\
- Do not repeat or alter any previously generated follow-up questions or intermediate answers.\\
- Conclude with the final answer using this exact format: 'So the final answer is: {final answer}' if no further questions are needed.\\
\\
Use the examples below to understand the expected structure, and follow this format without deviating from these instructions.\\
\\
Question: Who lived longer, Muhammad Ali or Alan Turing?\\
Are follow up questions needed here: Yes.\\
Follow up: How old was Muhammad Ali when he died?\\
Intermediate answer: Muhammad Ali was 74 years old when he died.\\
Follow up: How old was Alan Turing when he died?\\
Intermediate answer: Alan Turing was 41 years old when he died.\\
So the final answer is: Muhammad Ali.\\
\\
Question: When was the founder of craigslist born?\\
Are follow up questions needed here: Yes.\\
Follow up: Who was the founder of craigslist?\\
Intermediate answer: Craigslist was founded by Craig Newmark.\\
Follow up: When was Craig Newmark born?\\
Intermediate answer: Craig Newmark was born on December 6, 1952.\\
So the final answer is: December 6, 1952.\\
... 
\\
---\\
Now, **continue the response** using the following question and information provided below.\\
Only add follow-up questions if necessary to reach the final answer.\\
**Ensure all follow-up questions are optimized for search engine queries, making each question concise, direct, and easily searchable. Avoid modifying or repeating any existing content.**\\
---\\
\\
Question (ORIGINAL): \{question\}\\
Are follow up questions needed here: Yes.\\
\end{tcolorbox}
\caption{Question Decomposition prompt template.}
\label{appendix:decomposition prompt}
\end{figure*}

\begin{figure*}[t]
\noindent
\begin{tcolorbox}[title = {Retrieve-then-Read Prompt}] 
\#\# Question-Answering-in-Reference-Task \#\#\\
\\
Instruction:\\
- Use the references provided to answer the question as specifically and completely as possible.\\
- If the references do not directly answer the question, combine relevant information from multiple references to create a well-supported answer.\\
- When references are Null or insufficient, use your own knowledge to provide a clear and relevant answer.\\
- When a direct answer cannot be determined, list any information in the references that could be relevant or provide partial insights related to the question. Avoid responses such as 'I don’t know' or 'more information is needed.'\\
- Always prioritize specificity and relevance in your answer, providing helpful context or details that approach a complete answer.\\
\\
Reference [1]\\
Reference [2]\\
...\\
\\
Question: \{question\}\\
\end{tcolorbox}
\caption{Retrieve-then-Read prompt template.}
\label{appendix:generation prompt}
\end{figure*}

\begin{figure*}[t]
\noindent
\begin{tcolorbox}[title = {Self-Critique Prompt}] 
\#\# Critique-Task \#\# \\
Main Question: \{question\}\\
Below is a list of previously generated subquestions and their intermediate answers, created as part of a multi-step reasoning process to answer the main question.\\
Your task is to evaluate whether the information in the current subquestion is necessary and contributes incrementally towards solving the main question.\\
\\
Previously generated subquestions and answers:\\
 \{previous subquestions\}
\\
\\Current subquestion and answer candidate:\\
\{subquestion and intermediate answer \}\\
\\
Instruction:\\
- Step 1: Check for Redundancy. Check if the current subquestion or answer repeats information already provided in previous subquestions. If it does, return 'flag = False' as this information is redundant.\\
- Step 2: Assess Relevance. If the information is not a duplicate, analyze its relevance to the main question. Determine whether it provides new, relevant information that helps move closer to solving the main question, even if it only provides indirect context or background.\\
Note that information does not need to directly answer the main question to be considered relevant; it can also support understanding or provide necessary context. Mark it as 'flag = True'.\\
- Step 3: Based on your analysis, provide a final judgment in the following format:\\

**Final Judgment**: [flag = True or flag = False]\\

Examples:\\
\\
Main Question: “Who lived longer, Muhammad Ali or Alan Turing?”\\
• Follow up: “How old was Muhammad Ali when he died?” (Flag = True, relevant for lifespan comparison.)\\
• Follow up: “How old was Alan Turing when he died?” (Flag = True, completes lifespan comparison.)\\
• Redundant Example: “How old was Muhammad Ali when he passed?” (Flag = False, redundant with earlier subquestion.)\\
Main Question: “Are both the directors of Jaws and Casino Royale from the same country?”\\
• Follow up: “Who directed Jaws?” (Flag = True, needed for director identification.)\\
• Follow up: “Where is Steven Spielberg from?” (Flag = True, relevant to nationality check.)\\
• Irrelevant Example: “What is Steven Spielberg’s favorite genre?” (Flag = False, not relevant to nationality.)\\
\\
Reminder: Use “flag = True” for any subquestion that provides useful information or context toward solving the main question, even if indirectly. Set “flag = False” only if it is redundant or entirely irrelevant.
\end{tcolorbox}
\caption{Self-Critique prompt template.}
\label{appendix:critique prompt}
\end{figure*}

\subsubsection{Self-Decomposition Evaluation Prompt}
\label{appendix:gpt4decomposition}
In this paper, the evaluation of the question decomposition capability is conducted using GPT-4o
 with prompt as shown in Figure \ref{appendix:gpt4 decomposition prompt}. The analysis involves assessing and scoring the decomposition results of different iterations across multiple dimensions, ultimately leading to a comparative analysis of the two models. The dimensions of the analysis include:

\begin{itemize}
    \item \textbf{Conciseness}: Whether the decomposition avoids redundancy while ensuring comprehensiveness.
    \item \textbf{Rationality}: Whether the decomposed sub-problems are closely related to the original problem.
    \item \textbf{Sequencing}: Whether the decomposition of sub-problems follows a logical order and facilitates the problem-solving process.
    \item \textbf{Goal Orientation}: Whether the decomposition is clearly centered around addressing the main problem’s objective. Are the sub-problems closely aligned with the core goal of the main problem? Does it avoid redundant issues that deviate from the primary objective?
\end{itemize}
\begin{figure*}[htbp]
\noindent
\begin{tcolorbox}[title = {Self-Decomposition Evaluation Prompt}] 

You are given two problem decomposition results for the same complex problem. Your task is to compare these results from Conciseness, Rationality, Sequencing and Goal Orientation. Analyze the two decomposition results using the criteria above. Clearly explain which approach is more effective for solving the problem and why, while highlighting the strengths and weaknesses of each approach in detail. 
\\

\# Scoring Criteria: \\
- Score each dimension on a scale of 1-5, where: \\
  - 1: Poor \\
  - 2: Needs Improvement \\
  - 3: Average \\
  - 4: Good \\
  - 5: Excellent \\
\# The output follows the format below. Do not add any additional text:
\{ \\
    "Conciseness": \{ \\
        "Result 1 Score": X, \\
        "Result 2 Score": Y, \\
        "Explanation": "How effectively does each decomposition avoid unnecessary complexity while still addressing all relevant aspects of the problem? Is the explanation clear and straightforward?" \\
    \}, \\
    "Rationality": \{ \\
        "Result 1 Score": X, \\
        "Result 2 Score": Y, \\
        "Explanation": "Are the identified components logical and directly related to the problem? Do the solutions align well with the identified components?" \\
    \}, \\
    "Sequencing": \{ \\
        "Result 1 Score": X, \\
        "Result 2 Score": Y, \\
        "Explanation": "Is the order of steps or components logical and easy to follow? Does the sequence facilitate efficient problem-solving?" \\
    \}, \\
    "Goal Orientation": \{ \\
        "Result 1 Score": X, \\
        "Result 2 Score": Y, \\
        "Explanation": "Do the sub-questions stay aligned with the core goal of the main problem? Are there any redundant sub-questions that deviate from the primary objective?" \\
    \}, \\
    "Result": "Decomposition Results 1\ Decomposition Results 2\ Tie" \\
\} \\
\# Problem: \\
\{problem\}
\# Decomposition Results to Compare:

- Decomposition Results 1: \\
\{result1\}

- Decomposition Results 2: \\
\{result2\}
\# Output:
\end{tcolorbox}
\caption{GPT-4o decomposition prompt template.}
\label{appendix:gpt4 decomposition prompt}
\end{figure*}

\subsection{Experiment detail}
\label{experiment details}

\subsubsection{Implementation Details}
\label{implementation}
We conduct all experiments on a server equipped with four NVIDIA A800 80G GPUs. For the experimental setup, we use the following hyperparameters: learning rate of \(1 \times 10^{-4}\), batch size of 64,and cut-off length of 8192. Furthermore, for the weighting parameters $\alpha$, $\beta$, and$\gamma$ in the overall loss function, values of 1 are uniformly adopted in this research.

\subsubsection{Datasets}
The cold-start dataset $Q^0$ consists of 800 randomly sampled instances from the training sets of 2WikiMultiHopQA, HotpotQA, and MuSiQue, totaling 2,400 cold-start samples. Table \ref{tab:train_data} provides detailed information on the training datasets constructed during each round of self-exploration. The evaluation datasets we used 2WikiMultiHopQA, HotpotQA, and MuSiQue each contain 1,000 samples, Nature Questions, Web Questions, and TriviaQA each contain 400 samples.

\begin{table}[]
\centering
\begin{tabular}{lllll}
\hline
Datasets &$\mathcal{D}_d$    &$\mathcal{D}_r$  & $\mathcal{D}_c$  &  \\\hline
Round 1 & 3276 & 2501 & 3925 &  \\ 
Round 2 & 8309 & 6311 & 8074 &  \\ 
Round 3 & 4858 & 2106 & 2312 & \\
Round 4 & 6913 & 4759 & 5307 & \\ \hline
\end{tabular}
\caption{Number of samples accumulated in datasets $\mathcal{D}_d$, $\mathcal{D}_r$, and $\mathcal{D}_c$ after each round of self-exploration.}
\label{tab:train_data}
\end{table}

\subsubsection{Self-Critique Capability Experiments Details}
\label{appendix:self-criticism}
To demonstrate the improvement in the self-critique capability of the model across iterations, we sampled 300 instances from the generated $\mathcal{D}_c$ at each round and compared them with GPT-4o. The responses from GPT-4o were used as ground truth to calculate the self-critique accuracy of our model. In Table \ref{tab:self_critique_num}, we present the number of instances in each round’s self-critique capability evaluation that aligned with GPT-4o.

\begin{table}[]
\centering
\begin{tabular}{cccc}
\hline
       & 2WIKI & HotpotQA & MSQ \\ \hline
Round 1 &  223     &  194        &  180   \\
Round 2 &  218     &  199        &  228   \\
Round 3 &  239     &  232        &  238   \\
Round 4 &  242     &  238        &  234   \\
Total   &  300     &  300        &  300   \\ \hline
\end{tabular}
\caption{Number of instances in each round’s self-critique capability evaluation that aligned with GPT-4o.}
\label{tab:self_critique_num}
\end{table}

\begin{table*}[htbp]
\centering
\begin{tabular}{l|lllllllll}
\hline
        & \multicolumn{3}{c}{2WIKI}      & \multicolumn{3}{c}{HotpotQA} & \multicolumn{3}{c}{MSQ} \\
        & Acc            & F1    & EM    & Acc      & F1      & EM      & Acc    & F1     & EM    \\ \hline
Round 0 & 38.98          & 23.28 & 12.10 & 31.94    & 21.72   & 10.28   & 7.30   & 8.98   & 2.70  \\
Round 1 & 45.97          & 46.49 & 32.50 & 35.83    & \textbf{44.55}   & \textbf{32.50 }  & 10.54  & 18.65  & 8.92  \\
Round 2 & 46.24          & 46.33 & 30.83 & 36.11    & 41.58   & 30.00   & 11.89  & 16.19  & 8.92  \\
Round 3 & \textbf{47.31} & \textbf{47.10} &\textbf{34.41} & \textbf{38.33}    & 42.82   & 31.39   & \textbf{13.24}  & \textbf{20.70} & \textbf{10.81} \\ \hline
\end{tabular}
\caption{Performance of the Qwen2.5-7B model after four rounds of self-exploration on different datasets, showing improvements in accuracy, F1, and EM scores across 2WIKI, HotpotQA, and MSQ.}
\end{table*}

\begin{table*}[htbp]
\centering
\begin{tabular}{cccc}
\hline
Methods & Average Input Token & Average Output Tokens & Number of LLMs Calls \\ \hline
Naive LLM & 18   & 244  & 1  \\
CoT       & 24   & 284  & 1  \\
CoT-SC    & 240  & 2840 & 10 \\
GenRead   & 307  & 233  & 2  \\
Naive RAG & 440  & 104  & 1  \\
Self-Ask  & 782  & 172  & 2   \\
WebGLM    & 436  & 103  & 1  \\
Self-RAG  & 866  & 294  & 2  \\
RRR       & 496  & 86   & 2  \\
GenGround & 2449 & 167  & 5  \\
RISE      & 2881 & 192  & 5  \\ \hline
\end{tabular}
\caption{Token consumption comparison between RISE and other baseline methods, showing average input tokens, average output tokens, and the number of LLMs calls required for each approach. RISE demonstrates a higher input token consumption due to its multi-step reasoning, but maintains efficient reasoning performance.}
\end{table*}

\subsection{Additional Experiments}
\subsubsection{RISE Robustness}
To further verify the robustness of our experimental conclusions, we conducted additional experiments using the Qwen2.5-7B model. Specifically, we performed four rounds of self-exploration following the same experimental setup as in our original RISE framework. The results consistently demonstrate the effectiveness of RISE, with performance improvements observed across multiple datasets after each iteration. This confirms that RISE maintains strong generalization capabilities and stable performance even when applied to different large language models.

\subsubsection{Token Consumption Details}
In addition to performance evaluation, we analyzed the token consumption of RISE compared to other baseline methods. We measured both the average input token consumption and the average output token length, as well as the number of model calls required in each approach. The results reveal that while RISE consumes more input tokens due to its multi-step reasoning process, it achieves higher efficiency in output generation and overall reasoning effectiveness. This analysis highlights the trade-off between token usage and model performance, demonstrating that RISE achieves a balanced optimization in complex reasoning tasks.

\subsubsection{The necessity of multiple iterations}
To further validate the necessity and effectiveness of our multi-round training strategy, we conduct additional experiments comparing single-round and multi-round training setups on three benchmark datasets: 2Wiki, HotpotQA, and MSQ. The results are summarized in Table~\ref{tab:multi_round_training}.

These results demonstrate that multi-round training significantly improves accuracy (Acc) and exact match (EM) metrics across all datasets, highlighting the advantage of iterative self-exploration over static single-round training.  
Additionally, we compare joint training with alternating training strategies to clarify their differences. As shown in Table~\ref{tab:joint_vs_alternating}, joint training better preserves learned capabilities and achieves performance gains through synergistic task-chain interactions.

\begin{table*}[htbp]
    \centering
    \label{tab:multi_round_training}
    \begin{tabular}{lccccccccc}
        \toprule
        \multirow{2}{*}{Setup} & \multicolumn{3}{c}{2Wiki} & \multicolumn{3}{c}{Hotpot} & \multicolumn{3}{c}{MSQ} \\ 
        & Acc & F1 & EM & Acc & F1 & EM & Acc & F1 & EM \\ 
        \midrule
        Single-round & 41.10 & 43.32 & 30.60 & 34.10 & 43.28 & 29.50 & 11.10 & 17.78 & 8.60 \\
        Multi-round  & \textbf{49.40} & 43.01 & \textbf{32.70} & \textbf{40.50} & 42.00 & \textbf{30.50} & \textbf{21.70} & \textbf{22.87} & \textbf{11.70} \\
        \bottomrule
    \end{tabular}
    \caption{Comparison between single-round and multi-round training on three datasets.}
\end{table*}

\subsubsection{Reliability Analysis of Self-Evaluation Mechanism}
We further investigate the reliability of the model's self-critique mechanism by evaluating the alignment between self-assessment and actual correctness. Table~\ref{tab:self_evaluation_reliability} presents the distribution of cases where the model's self-judgment matches or mismatches the ground truth for both the baseline GPT-4o and our method.

These results confirm a positive correlation between the model's self-assessment and actual answer correctness, demonstrating the effectiveness and reliability of the proposed self-evaluation mechanism.


\begin{table*}[htbp]
    \centering
    \label{tab:self_evaluation_reliability}
    \resizebox{\textwidth}{!}{
    \begin{tabular}{lcccc}
        \toprule
        Model & Correct and judged Correct & Correct but judged Wrong & Wrong but judged Correct & Wrong and judged Wrong \\
        \midrule
        GPT-4o & 32.58\% & 9.71\% & 11.21\% & 46.50\% \\
        Ours   & 28.65\% & 7.58\% & 23.43\% & 40.33\% \\
        \bottomrule
    \end{tabular}
    }
    \caption{Alignment between model self-judgment and ground truth correctness on initial question set \(Q_0\).}
\end{table*}

\subsection{Supply Metrics of Main Results}
\label{sec:appendix-metrics}
This section provides additional details to supplement the main results, including comprehensive Exact Match (EM) and F1 scores across six QA datasets: 2WikiMultiHopQA, HotpotQA, MuSiQue, Natural Questions, Web Questions, and TriviaQA. We compare RISE (Ours) with both prompting-based and fine-tuning-based methods, under settings with and without retrieval. The results offer a deeper understanding of RISE’s performance, highlighting its consistent improvements over baseline models.

\begin{table*}[]
\centering
\scalebox{0.85} {
\begin{tabular}
{p{4.5cm}p{2.5cm}|p{1.35cm}p{1.35cm}p{1.35cm}p{1.35cm}p{1.35cm}p{1.35cm}p{1.35cm}llcccccc}
\hline
\multirow{2}{*}{\textbf{Method}} &
  \multirow{2}{*}{\textbf{Model}} &
  \multicolumn{3}{c|}{\textbf{MHQA}} &
  \multicolumn{3}{c}{\textbf{SHQA}} \\
   & &
  \textbf{2WIKI} &
  \textbf{Hotpot} &
  \textbf{MSQ} &
  \multicolumn{1}{c}{\textbf{NQ}} &
  \multicolumn{1}{c}{\textbf{WebQ}} &
  \multicolumn{1}{c}{\textbf{Trival}} \\ \hline
\rowcolor[gray]{0.95} \multicolumn{8}{c}{\textit{w/o retrieval}} \\ 
\rowcolor{green!10} Naive LLM                                                                       & LLaMA-3.1-8B    & 0.00  & 0.30  & 0.00  & 0.25  &  0.00  & 2.25  \\
\rowcolor{green!10}                                                                                 & GPT-3.5-turbo   & 0.50  & 5.50  & 0.20  & 2.50  &  0.75  & 26.75 \\
\rowcolor{green!10} CoT~\cite{wei2023chainofthoughtpromptingelicitsreasoning}                       & LLaMA-3.1-8B    & 0.00  & 0.00  & 0.00  & 0.00  &  0.00  & 0.25  \\
\rowcolor{green!10} CoT-SC\textsuperscript{*}~\cite{wang2023selfconsistencyimproveschainthought}    & GPT-3.5-turbo   & 18.40 & 31.90 & 9.60  & 30.75 &  22.75 & 71.75 \\
\rowcolor{green!10} GenRead~\cite{yu2023generateretrievelargelanguage}                              & LLaMA-3.1-8B    & 17.00 & 25.60 & 7.60  & 35.75 & 25.50  & 61.75 \\
\rowcolor[gray]{0.95} \multicolumn{8}{c}{\textit{w retrieval}} \\ 
\rowcolor{green!10} Naive RAG                                                                       & LLaMA-3.1-8B    & 0.30  & 1.70  & 0.20  & 0.00  & 0.00   & 2.25  \\
\rowcolor{green!10} Self-Ask~\cite{press2023measuringnarrowingcompositionalitygap}                  & LLaMA-3.1-8B    & 17.00 & 25.60 & 9.60  & 28.50 & 21.50  & 58.75 \\
\rowcolor{yellow!10} WebGLM~\cite{liu2023webglm}                                                    & LLaMA-3.1-8B    & 0.00  & 0.30  & 0.00  & 0.00  & 0.00   & 0.00  \\
\rowcolor{yellow!10} Self-RAG\textsuperscript{*}~\cite{asai2023selfrag}                             & LLaMA2-7B       & 2.90  & 3.90  & 0.80  & 0.00  & 0.00   & 2.00  \\
\rowcolor{yellow!10}                                                                                & LLaMA2-13B      & 3.40  & 2.50  & 0.40  & 0.00  & 0.00   & 4.50  \\
\rowcolor{yellow!10} RRR~\cite{ma-etal-2023-query}                                                  & LLaMA-3.1-8B    & 0.00  & 0.40  & 0.00  & 0.75  & 0.75   & 2.75  \\
\rowcolor{green!10}                                                                                 & GPT-3.5-turbo   & 3.20  & 2.20  & 0.30  & 2.50  & 2.00   & 20.00 \\
\rowcolor{yellow!10} GenGround~\cite{shi2024generatethengroundretrievalaugmentedgenerationmultihop} & LLaMA-3.1-8B    & 23.50 & 24.30 & 10.20 & 20.50 & 18.50  & 60.25 \\ \hline
\rowcolor{yellow!10} \textbf{RISE(Ours)}                                                            & LLaMA-3.1-8B    & 32.70 & 30.50 & 11.70 & 28.50 & 19.25  & 59.50 \\ \hline
\end{tabular}
}
\caption{\textbf{\textit{\textcolor{red}{EM}}} metrics of RISE with other methods on the 2WikiMultiHopQA, HotpotQA, MuSiQue, Natural Questions, Web Questions and TriviaQA. Methods marked with asterisk (*) involve specific considerations: CoT-SC uses GPT-3.5 due to LLaMA-3.1’s limitations in adhering to instructions, and Self-RAG employs publicly released model weights because its dataset is unavailable. All other methods are reproduced with LLaMA-3.1-8B. \textbf{\textcolor{green!20}{\rule{3mm}{3mm}}} represent \textit{Prompting-based Methods}, while \textbf{\textcolor{yellow!20}{\rule{3mm}{3mm}}} represent \textit{Finetuning-based Methods}.}
\end{table*}

\begin{table*}[]
\centering
\scalebox{0.85} {
\begin{tabular}
{p{4.5cm}p{2.5cm}|p{1.35cm}p{1.35cm}p{1.35cm}|p{1.35cm}p{1.35cm}p{1.35cm}p{1.35cm}llcccccc}
\hline
\multirow{2}{*}{\textbf{Method}} &
  \multirow{2}{*}{\textbf{Model}} &
  \multicolumn{3}{c|}{\textbf{MHQA}} &
  \multicolumn{3}{c}{\textbf{SHQA}} \\
   & &
  \textbf{2WIKI} &
  \textbf{Hotpot} &
  \textbf{MSQ} &
  \multicolumn{1}{c}{\textbf{NQ}} &
  \multicolumn{1}{c}{\textbf{WebQ}} &
  \multicolumn{1}{c}{\textbf{Trival}} \\ \hline
\rowcolor[gray]{0.95} \multicolumn{8}{c}{\textit{w/o retrieval}} \\ 
\rowcolor{green!10} Naive LLM                                                                       & LLaMA-3.1-8B    & 7.99  & 5.49  & 2.67  & 4.09  & 5.46   & 17.55 \\
\rowcolor{green!10}                                                                                 & GPT-3.5-turbo   & 13.38 & 17.66 & 8.10  & 16.53 & 19.14  & 44.11 \\
\rowcolor{green!10} CoT~\cite{wei2023chainofthoughtpromptingelicitsreasoning}                       & LLaMA-3.1-8B    & 2.95  & 2.31  & 1.53  & 1.77  & 2.19   & 4.93 \\
\rowcolor{green!10} CoT-SC\textsuperscript{*}~\cite{wang2023selfconsistencyimproveschainthought}    & GPT-3.5-turbo   & 24.31 & 39.59 & 15.36 & 40.98 & 36.54  & 80.65 \\
\rowcolor{green!10} GenRead~\cite{yu2023generateretrievelargelanguage}                              & LLaMA-3.1-8B    & 22.39 & 34.51 & 15.26 & 47.90 & 42.55  & 69.59 \\
\rowcolor[gray]{0.95} \multicolumn{8}{c}{\textit{w retrieval}} \\ 
\rowcolor{green!10} Naive RAG                                                                       & LLaMA-3.1-8B    & 6.39  & 9.60  & 4.02  & 9.02  & 9.72   & 19.37 \\
\rowcolor{green!10} Self-Ask~\cite{press2023measuringnarrowingcompositionalitygap}                  & LLaMA-3.1-8B    & 23.42 & 36.14 & 19.67 & 40.35 & 39.61  & 67.02 \\
\rowcolor{yellow!10} WebGLM~\cite{liu2023webglm}                                                    & LLaMA-3.1-8B    & 8.27  & 6.03  & 3.76  & 5.57  & 6.68   & 9.30 \\
\rowcolor{yellow!10} Self-RAG\textsuperscript{*}~\cite{asai2023selfrag}                             & LLaMA2-7B       & 17.38 & 15.44 & 6.02  & 25.56 & 21.45  & 11.34 \\
\rowcolor{yellow!10}                                                                                & LLaMA2-13B      & 14.82 & 13.41 & 6.70  & 15.65 & 9.34   & 15.34 \\
\rowcolor{yellow!10} RRR~\cite{ma-etal-2023-query}                                                  & LLaMA-3.1-8B    & 5.35  & 3.21  & 1.48  & 4.46  & 6.68   & 9.56 \\
\rowcolor{green!10}                                                                                 & GPT-3.5-turbo   & 13.70 & 16.76 & 6.08  & 22.80 & 23.16  & 43.68 \\
\rowcolor{yellow!10} GenGround~\cite{shi2024generatethengroundretrievalaugmentedgenerationmultihop} & LLaMA-3.1-8B    & 34.33 & 34.53 & 19.81 & 36.29 & 28.93  & 67.20 \\ \hline
\rowcolor{yellow!10} \textbf{RISE(Ours)}                                                            & LLaMA-3.1-8B    & 43.01 & 42.00 & 22.87 & 42.97 & 36.22  & 72.70 \\ \hline
\end{tabular}
}
\caption{\textbf{\textit{\textcolor{red}{F1}}} metrics of RISE with other methods on the 2WikiMultiHopQA, HotpotQA, MuSiQue, Natural Questions, Web Questions and TriviaQA. Methods marked with asterisk (*) involve specific considerations: CoT-SC uses GPT-3.5 due to LLaMA-3.1’s limitations in adhering to instructions, and Self-RAG employs publicly released model weights because its dataset is unavailable. All other methods are reproduced with LLaMA-3.1-8B. \textbf{\textcolor{green!20}{\rule{3mm}{3mm}}} represent \textit{Prompting-based Methods}, while \textbf{\textcolor{yellow!20}{\rule{3mm}{3mm}}} represent \textit{Finetuning-based Methods}.}
\end{table*}

\begin{figure*}[t]
\noindent
\begin{tcolorbox}[title = {Multi-Hop Question Generation Prompt}] 
\#Multi-Hop-Question-Generation-in-\{Task\}\#\\
\\
Instruction:\\
- You are an AI assistant tasked with generating multi-hop questions similar to those in the \{task\} dataset.\\
- These questions require combining multiple pieces of information to reach the answer.\\
- Typically, these questions involve indirect references and nested relationships.\\
\\
Examples from the \{task\} dataset:\\
Example 1: Question: [Insert Example Here]\\
Example 2: Question: [Insert Example Here]\\
Example 3: Question: [Insert Example Here]\\
Example 4: Question: [Insert Example Here]\\
Example 5: Question: [Insert Example Here]\\
Example 6: Question: [Insert Example Here]\\
Example 7: Question: [Insert Example Here]\\
Example 8: Question: [Insert Example Here]\\
\\
Please generate a **new and unique multi-hop question** that meets the following criteria:\\
- **Requires reasoning across multiple facts or entities**.\\
- **Asks for only one piece of information or answer** without multiple sub-questions.\\
- **Is a single, coherent question** that requires multi-step reasoning to answer.\\
- **Includes indirect references or nested relationships**.\\
- **Matches the complexity and structure** of questions in the \{task\} dataset.\\
- **Is concise** (one sentence) and clearly worded.\\
- **Covers new topics** or involves different entities from the examples above.\\
- **Avoids being a duplicate** of the examples above.\\
- **Avoids using conjunctions like 'and', 'or', or commas** to ask more than one thing.\\
\\
Your response should **only contain the generated question** without any extra text and follow the format:\\
Question: \{question\}\\
\end{tcolorbox}
\caption{Multi-hop question generation prompt template.}
\label{appendix:multi-hop-prompt}
\end{figure*}

\subsection{Case Study}
To further illustrate the design of self-exploration process, we present representative case study. Each task is marked with a task-specific prefix, and follows a carefully curated instruction template.(Shown in Figure~\ref{fig:case_study_qc_i}, \ref{fig:case_study_qc_o}, \ref{fig:case_study_rtr_i}, \ref{fig:case_study_rtr_o}, \ref{fig:case_study_sc_i}, \ref{fig:case_study_sc_o})

\begin{figure*}[ht]
\centering
\begin{tcolorbox}[width=0.95\textwidth, colback=gray!3, colframe=black!50, 
title=\textbf{Case Study: Question Decomposition Input}]
\small
\#\# Question-Decomposition-Task \#\# \\

Please answer the following questions according to the given format. Strictly follow each format specification, as this will ensure consistency and clarity in your response.

- Only add follow-up questions if additional details are needed to arrive at the final answer.\\
- For each follow-up question, use exactly this format: 'Follow up: {question}'\\
- Ensure each follow-up question is direct and structured to be easily searchable, focusing on key information for efficient search engine retrieval.\\
- For each answer to a follow-up question, use exactly this format: 'Intermediate answer: {answer}'\\
- Do not repeat or alter any previously generated follow-up questions or intermediate answers.\\
- Conclude with the final answer using this exact format: 'So the final answer is: {final answer}' if no further questions are needed.\\
\\
Use the examples below to understand the expected structure, and follow this format without deviating from these instructions.\\

\vspace{0.8em}
\noindent\texttt{Question: Who lived longer, Muhammad Ali or Alan Turing?} \\
\texttt{Are follow up questions needed here: Yes.} \\
\texttt{Follow up: How old was Muhammad Ali when he died?} \\
\texttt{Intermediate answer: Muhammad Ali was 74 years old when he died.} \\
\texttt{Follow up: How old was Alan Turing when he died?} \\
\texttt{Intermediate answer: Alan Turing was 41 years old when he died.} \\
\texttt{So the final answer is: Muhammad Ali.}
...

\vspace{0.8em}
\noindent\texttt{---}

Now, continue the response using the following question and information provided below. \\
Only add follow-up questions if necessary to reach the final answer. \\
Ensure all follow-up questions are optimized for search engine queries, making each question concise, direct, and easily searchable. Avoid modifying or repeating any existing content.

\vspace{0.8em}
\noindent\texttt{---}

\texttt{Question: Are the directors of both films Inter Nos and La Bandera (Film) from the same country?} \\
\texttt{Are follow up questions needed here:}
\end{tcolorbox}
\caption{Example of Question Decomposition task input.}
\label{fig:case_study_qc_i}
\end{figure*}

\begin{figure*}[ht]
\centering
\begin{tcolorbox}[width=0.95\textwidth, colback=gray!3, colframe=black!50, 
title=\textbf{Case Study: Question Decomposition Output}]
\small{
\texttt{Yes.} \\ 
\texttt{Follow up: Who is the director of Inter Nos?}
}
\end{tcolorbox}
\caption{Example of Question Decomposition task output.}
\label{fig:case_study_qc_o}
\end{figure*}

\begin{figure*}[ht]
\centering
\begin{tcolorbox}[width=0.95\textwidth, colback=gray!3, colframe=black!50, 
title=\textbf{Case Study: Retrieve-then-Read Input}]
\small
\#\# Question-Answering-in-Reference-Task \#\# \\

Instruction:
- Use the references provided to answer the question as specifically and completely as possible.
- If the references do not directly answer the question, combine relevant information from multiple references to create a well-supported answer.
- When references are Null or insufficient, use your own knowledge to provide a clear and relevant answer.
- When a direct answer cannot be determined, list any information in the references that could be relevant or provide partial insights related to the question. Avoid responses such as 'I don’t know' or 'more information is needed.'
- Always prioritize specificity and relevance in your answer, providing helpful context or details that approach a complete answer.

\vspace{0.8em}
\noindent\texttt{Reference [1]: Congratulations to the four Directors of Inter Nos... \\
Reference [2]: ...\\
...(More retrieve contents)}

\vspace{0.8em}
\noindent\texttt{---}

\texttt{Question: Who is the director of Inter Nos?} \\
\end{tcolorbox}
\caption{Example of Retrieve-then-Read task input.}
\label{fig:case_study_rtr_i}
\end{figure*}

\begin{figure*}[ht]
\centering
\begin{tcolorbox}[width=0.95\textwidth, colback=gray!3, colframe=black!50, 
title=\textbf{Case Study: Retrieve-then-Read Output}]
\small{
\texttt{According to the references provided, the directors of Inter Nos include President Juan Carlos Martínez. Additionally, Cynthia Vaughn has been named Associate Editor of Inter Nos, and she will edit the publication's “Independent Voices” section. It is not specified in the references provided who the other directors of Inter Nos are.}
}
\end{tcolorbox}
\caption{Example of Retrieve-then-Read task output.}
\label{fig:case_study_rtr_o}
\end{figure*}

\begin{figure*}[ht]
\centering
\begin{tcolorbox}[width=0.95\textwidth, colback=gray!3, colframe=black!50, 
title=\textbf{Case Study: Self-Critique Input}]
\small
\#\# Critique-Task \#\# \\
Main Question: Are the directors of both films Inter Nos and La Bandera (Film) from the same country? \\

Below is a list of previously generated subquestions and their intermediate answers, created as part of a multi-step reasoning process to answer the main question.
Your task is to evaluate whether the information in the current subquestion is necessary and contributes incrementally towards solving the main question.\\

Previously generated subquestions and answers:
Null \\

\vspace{0.8em}
\noindent\texttt{Current subquestion and answer candidate: \\
Follow up: Who is the director of Inter Nos? \\
Intermediate answer: According to the references provided, the directors of Inter Nos  include President Juan Carlos Martínez. Additionally, Cynthia Vaughn has been named Associate Editor of Inter Nos, and she will edit the publication's “Independent Voices” section. It is not specified in the references provided who the other directors of Inter Nos are.} \\

Instruction: \\
- Step 1: Check for Redundancy. Check if the current subquestion or answer repeats information already provided in previous subquestions. If it does, return 'flag = False' as this information is redundant. \\
- Step 2: Assess Relevance. If the information is not a duplicate, analyze its relevance to the main question. Determine whether it provides new, relevant information that helps move closer to solving the main question, even if it only provides indirect context or background.Note that information does not need to directly answer the main question to be considered relevant; it can also support understanding or provide necessary context. Mark it as “flag = True. \\
- Step 3: Based on your analysis, provide a final judgment in the following format: \\
**Final Judgment**: [flag = True or flag = False] \\

Examples: \\
\texttt{Main Question: “Who lived longer, Muhammad Ali or Alan Turing?” \\
 • Follow up: “How old was Muhammad Ali when he died?” (Flag = True, relevant for lifespan comparison.) \\
 • Follow up: “How old was Alan Turing when he died?” (Flag = True, completes lifespan comparison.) \\
 • Redundant Example: “How old was Muhammad Ali when he passed?” (Flag = False, redundant with earlier subquestion.) \\}

\texttt{Main Question: “Are both the directors of Jaws and Casino Royale from the same country?” \\
 • Follow up: “Who directed Jaws?” (Flag = True, needed for director identification.) \\
 • Follow up: “Where is Steven Spielberg from?” (Flag = True, relevant to nationality check.) \\ 
 • Irrelevant Example: “What is Steven Spielberg’s favorite genre?” (Flag = False, not relevant to nationality.)} \\

Reminder: \\
Use “flag = True” for any subquestion that provides useful information or context toward solving the main question, even if indirectly. Set “flag = False” only if it is redundant or entirely irrelevant.
\end{tcolorbox}
\caption{Example of Self-Critique task input.}
\label{fig:case_study_sc_i}
\end{figure*}

\begin{figure*}[ht]
\centering
\begin{tcolorbox}[width=0.95\textwidth, colback=gray!3, colframe=black!50, 
title=\textbf{Case Study: Self-Critique Output}]
\small{
\texttt{flag = True.}
}
\end{tcolorbox}
\caption{Example of Self-Critique task output.}
\label{fig:case_study_sc_o}
\end{figure*}

\end{document}